\documentclass[sigconf,natbib=false]{acmart}

\AtBeginDocument{%
  }

\copyrightyear{2024}
\acmYear{2024}
\setcopyright{acmlicensed}
\acmConference[ICMR '24] {Proceedings of the 2024 International Conference on Multimedia Retrieval}{June 10--14, 2024}{Phuket, Thailand.}
\acmBooktitle{Proceedings of the 2024 International Conference on Multimedia Retrieval (ICMR '24), June 10--14, 2024, Phuket, Thailand}
\acmPrice{15.00}
\acmISBN{979-8-4007-0619-6/24/06}
\acmDOI{10.1145/3652583.3658095}

\RequirePackage[
  datamodel=acmdatamodel,
  style=acmnumeric,
  ]{biblatex}

\usepackage{multirow}
\usepackage{balance} 
\begin{document}

\title{Calibration \& Reconstruction: Deep Integrated Language for Referring Image Segmentation}

\author{Yichen Yan}
\email{yanyichen2021@ia.ac.cn}

\affiliation{
  \institution{$^{1}$ Institute of Automation, Chinese Academy of Sciences, $^{2}$School of Artificial Intelligence, University of Chinese Academy of Sciences
} 
  \city{Beijing}
  \country{China}
}

\author{Xingjian He}
\email{xingjian.he@nlpr.ia.ac.cn}
\affiliation{
  \institution{Institute of Automation, Chinese Academy of Sciences} 
  \city{Beijing}
  \country{China}
}

\author{Sihan Chen}
\email{Sihan.Chen@nlpr.ia.ac.cn}
\affiliation{
  \institution{School of Artificial Intelligence, University of Chinese Academy of Sciences} 
  \city{Beijing}
  \country{China}
}

\author{Jing Liu}
\email{jliu@nlpr.ia.ac.cn}
\authornote{corresponding author}
\affiliation{
  \institution{$^{1}$ Institute of Automation, Chinese Academy of Sciences, $^{2}$School of Artificial Intelligence, University of Chinese Academy of Sciences
} 
  \city{Beijing}
  \country{China}
}

\settopmatter{printacmref=true} 

\begin{abstract}
Referring image segmentation aims to segment an object referred to by natural language expression from an image. The primary challenge lies in the efficient propagation of fine-grained semantic information from textual features to visual features.  Many recent works utilize a Transformer to address this challenge.
However, conventional transformer decoders can distort linguistic information with deeper layers, leading to suboptimal results.    In this paper, we introduce CRFormer, a model that iteratively calibrates multi-modal features in the transformer decoder.  We start by generating language queries using vision features, emphasizing different aspects of the input language. Then, we propose a novel Calibration Decoder (CDec) wherein the multi-modal features can iteratively calibrated by the input language features. In the Calibration Decoder, we use the output of each decoder layer and the original language features to generate new queries for continuous calibration, which gradually updates the language features.  Based on CDec, we introduce a Language Reconstruction Module and a reconstruction loss. This module leverages queries from the final layer of the decoder to reconstruct the input language and compute the reconstruction loss. This can further prevent the language information from being lost or distorted. Our experiments consistently show the superior performance of our approach across RefCOCO, RefCOCO+, and G-Ref datasets compared to state-of-the-art methods.

\end{abstract}


\maketitle

\begin{figure}[t]
  \centering
  \includegraphics[width=\linewidth]{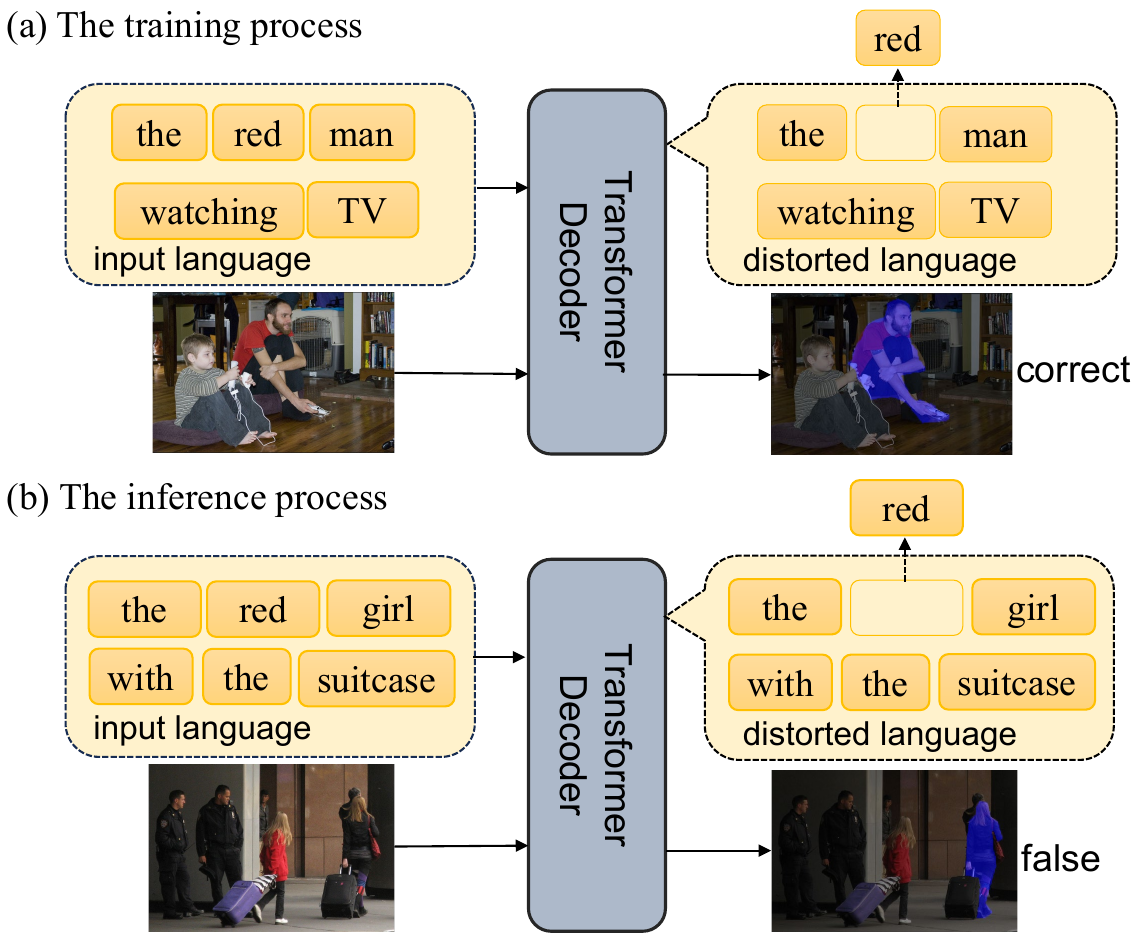}
  \caption{Distortion of language information propagation in the transformer decoder. (a) In the training process, the distortion of language information can still obtain the correct result in the image which has only two prominent targets by coincidence, but the traditional cross-entropy loss can not reflect this coincidence. (b) During the inference stage, the distortion of language information may result in inaccurate segmentation masks, especially in images featuring multiple targets, such as two girls with suitcases.}
\end{figure}

\section{Introduction}
Referring image segmentation (RIS) \cite{hu2016segmentation,liu2017recurrent,li2018referring}, which aims to segment an object referred to by natural language expression from an image. This task has numerous potential applications, including interactive image editing and human-object interaction \cite{wang2019reinforced}. One of the biggest challenges is how to propagate fine-grained semantic information from textual features to visual features. 

Recently, many methods utilize Transformer to adaptively facilitate fine-grained semantic information. Leveraging the Transformer's adept global perspective notably enhances the accuracy.  Nonetheless, this approach presents two primary drawbacks:

(a) As the layer count in the conventional Transformer decoder increases, the vision features of the query continuously fuse with language features, iteratively forming new multi-modal features. Concurrently, the language features of the key and value remain unaltered, thereby impeding the propagation process. 

(b) In the standard Transformer decoder, as the depth of the decoder layers increases, there is a potential loss or distortion of crucial language information. Moreover, the loss function of previous RIS methods solely comprises the cross-entropy loss based on segmentation results, akin to the loss function in traditional image segmentation. This loss function merely considers segmentation outcomes at the visual level and cannot gauge the degree of language-level distortion. 

We argue that the language representation should be updated gradually during the decoder's operation. As illustrated in Figure 1 (a), in the training process, the input language is \emph{"the red man watching TV"}. As the transformer grows in layers, the word \emph{"red"} becomes distorted. However, with only a man and a child as two obvious targets in the image, the network could accidentally obtain the result based on the distorted language \emph{"man"} and \emph{"watching TV"} or its past experience without understanding the linguistic semantic information. The traditional mask cross-entropy function cannot measure this coincidence, so the network can easily learn such kind of data bias and always output based on the distorted language.  Therefore, for some training samples, if the network happens to “guess” the correct target even if the language information has been lost during the propagation, these training samples may not properly contribute to the training process, or even impede the framework. In the inference process, this language distortion can lead to false results.  As shown in Figure 1 (b), where the crucial word "red" is absent, and the image presents multiple targets, including two women holding suitcases, the model is prone to generating inaccurate results. 

To address this problem, initially, we generate multiple language queries representing various emphases and detailed semantic information. These queries, enriched with detailed information, help mitigate natural distortion during the decoder propagation process. This stands in contrast to original language features that only contain a single modality of information. Meanwhile, we believe that  language propagation is a gradual process, which means the language features should also be constantly updated with the process of propagation. So we design a novel Calibration Decoder which can continuously calibrate the language information by generating new language queries in each decoder layer. In order to evaluate the distortion degree of language information after continuous correction, we also design a Language Reconstruction Module and language reconstruction loss. This module reconstructs the input language features with language queries from the last layer of the decoder. Finally, we use reconstruction loss and cross-entropy loss as the final loss of our model to optimize the network. In summary, our main contributions are listed as follows:
\begin{itemize}
    \item We propose a novel framework named CRFormer which  deep integrates the language representation to address the problem of language information distortion during the semantic propagation process in referring image segmentation.
    \item We design a Calibration Decoder (CDec) to gradually update the language features, which continuously calibrates the language features to avoid distortion.  We also design a novel language reconstruction loss to supervise the language information propagation.
    \item We achieve new state-of-the-art results on three datasets for referring image segmentation, demonstrating the effectiveness of our proposed method.
\end{itemize}
\section{Related Work}
\textbf{Referring Image Segmentation.} Referring image segmentation aims at segmenting a specific region in an image by comprehending a given natural language expression \cite{hu2016segmentation}. The main challenge is effectively propagating fine-grained semantic information from textual features to visual features. In early works \cite{hu2016segmentation,li2018referring,liu2017recurrent}, visual and textual features were initially extracted using Convolutional Neural Networks (CNN) and Long Short-Term Memory (LSTM), respectively. Subsequently, these features were directly concatenated to derive the final segmentation results. With increasing interest in attention mechanisms, a series of works\cite{hu2020bi,ding2021vision,ding2022vlt,liu2023multi,wu2023towards} have emerged to leverage cross-modal attention for the facilitation of semantic information propagation in a global field.  However, in the previous methods, linguistic information contained within the multi-modal features has a tendency to become distorted, resulting in suboptimal propagation. Moreover, they only utilize cross-entropy loss to optimize the modal which can not evaluate the distortion of language information. In contrast, we propose a series of approaches to address this problem and design a language reconstruction loss to supervise semantic information propagation directly.

\textbf{Transformer.} Transformer \cite{vaswani2017attention} is first designed for the Natural Language Processing (NLP) task. Thanks to its strong global relationship modeling ability, it was utilized in the Computer Vision (CV) task recently and achieved good performance in many tasks \cite{xie2021segformer,dosovitskiy2020image,carion2020end}. In the vision-language tasks, transformer architectures have achieved great success in many tasks \cite{kim2021vilt,lu2019vilbert,ramesh2021zero,huang2020pixel,chen2021history}. For referring image segmentation, numerous methods utilize transformers to adaptively propagate fine-grained semantic information from textual features to visual features.  However, in a traditional transformer decoder, language information may be gradually distorted or lost as the number of layers deepens, resulting in a suboptimal semantic propagation. Conversely, we propose a novel Calibration Decoder (CDec), which can continuously calibrate the language features to prevent semantic information distortion.

\begin{figure*}[t]
  \centering
  \includegraphics[width=\linewidth]{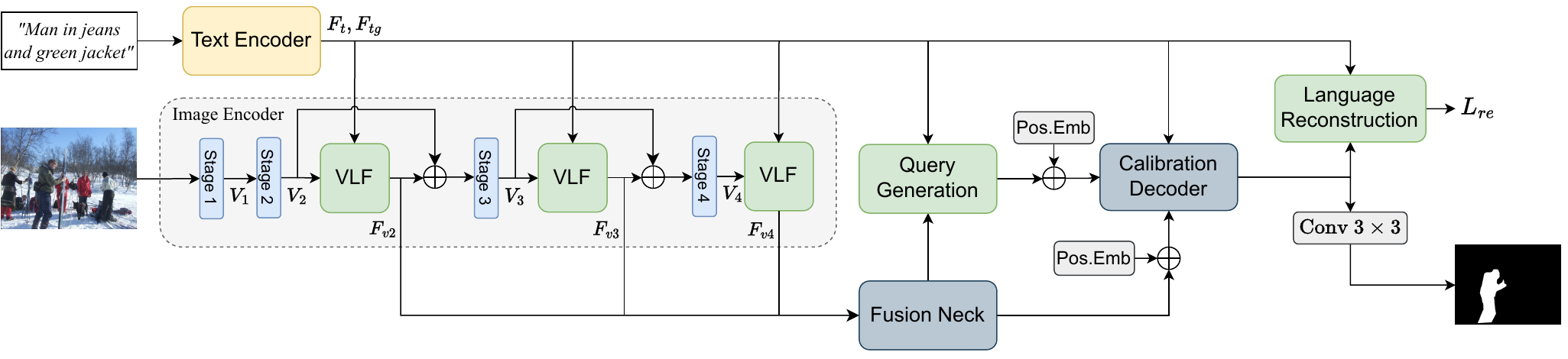}
  \caption{The overview of our proposed CRFormer. Our method mainly consists of a text encoder, an image encoder, a Query Generation Module, a Calibration Decoder, and a Language Reconstruction Module.}
\end{figure*}

\section{Methodology}

As illustrated in Figure 2, our proposed framework facilitates semantic information propagation and prevents language distortion. Firstly, the framework takes an image and a language expression as input. We employ Swin Transformer \cite{liu2021swin} and BERT\cite{devlin2018bert} to extract image and text features, respectively. Then we utilize a Query Generation Module to generate a series of language queries that represent different comprehension of the input sentence. These queries incorporate more detailed semantic information, thereby mitigating the likelihood of distortion. Then we propose a novel Calibration Decoder (CDec) to continuously integrate language semantic information with the vision features, which achieves effective cross-modal propagation. Then we design a Language Reconstruction Module to supervise the language propagation process. 

\begin{figure}[htbp]
  \centering
  \includegraphics[width=\linewidth]{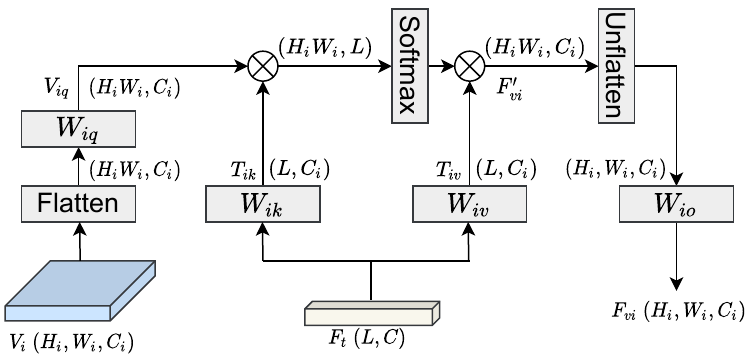}
  \caption{The details of Vision-Language Fusion Module. }
  \vspace{0.0mm}
\end{figure}

\subsection{Image and Text Feature Extraction}

\textbf{Text Encoder}. For a given language expression $\emph{T} \in \mathbb{R}^{\emph{L}}$, we utilize BERT\cite{devlin2018bert} to obtain text features $\emph{$F_t$} \in \mathbb{R}^{\emph{L}\,\times\,\emph{C}}$.  Additionally, we use the highest layer's activations of the BERT at the \texttt{[CLS]} token as the global feature for the entire language expression. This feature is linearly transformed and denoted as $F_{tg} \in \mathbb{R}^{1 \times C}$. Here, $C$ represents the feature dimension, while $L$ is the length of the language expression.

\textbf{Image Encoder}. Previous work\cite{yang2022lavt} on RIS has indicated that achieving better results is possible by fusing multi-modal features at an earlier stage in the encoding process. Inspired by this, we design an image encoder that fuses the vision and language features in the early phase. For a given image $I \in \mathbb{R}^{H \times W \times 3}$, we use the Swin Transformer \cite{liu2021swin} encoder. There are 4 stages in total and we denote the features as $\left\{V_{i}\right\}_{i=1}^{4}\in \mathbb{R}^{H_{i} \times W_{i} \times C_i}$. Here, $H_i$, $W_i$, and $C_i$ are the height, width, and number of channels of $\mathbf{x}_i$. In our model, unlike the original Swin Transformer, we integrate language features directly into the vision encoder. This fusion of language features occurs through our introduced Vision-Language Fusion module (VLF), which is positioned between each stage of the vision processing. Specifically, we denote VLF modules as $\{\delta_i|i \in \{2,3,4\}\}$ and four stages in the Swin Transformer as $\{v_i|i \in \{1,2,3,4\}\}$. The multi-modal calculation process in the encoding stage can be formed as follows:

\begin{equation}
\begin{aligned}
&V_1=v_1(I) \\
&V_2=v_2\left(V_1\right) \\
&F_{vi}=\delta_i\left(V_i, F_t\right), i \in\{2,3,4\} \\
&V_i=v_i(V_{i-1}+\sigma\left(F_{vi}\right)), i \in\{3,4\},
\end{aligned}
\end{equation}

Here, $\{F_{vi}\}_{i=2}^{4}\in \mathbb{R}^{H_{i} \times W_{i} \times C_i }$ denote the fused features and $\sigma$ denotes ReLu function. After each stage in the vision transformer
(except stage 1), a VLF fuses vision and language features together to get a multi-modal
feature $F_{vi}$. Then the multi-modal feature is fed as input to the next stage. Finally, we get three hierarchical multi-modal features $\{F_{vi}\}_{i=2}^{4}$. The specific details of the VLF module are in the next paragraph.

\textbf{Vision-Language Fusion Module}. 
As illustrated in  Figure 1, we employ cross-attention to fuse the language features $F_t$ and hierarchical features $\left\{V_{i}\right\}_{i=2}^{4}$. The VLF module takes $F_t$ and $V_i$ as input, then we flatten $v_i$ and utilize a linear projection $W_{iq}$ to transform it into $V_{iq}$. The language features $F_t$ are linear projected by $W_{ik}$ and $W_{iv}$ into $T_{ik}$ and $T_{iv}$. Then we calculate the attention matrix by $V_{iq}$ and $T_{ik}$. The output $F_{vi}$ is computed through $T_{iv}$ and attention matrix. The process is formulated as follows:
\begin{equation}
\begin{gathered}
    V_{i q}=W_{i q}\left( Flatten \left(v_i\right)\right), \\
T_{i k}=W_{i k} F_t, \\
T_{i v}=W_{i v} F_t, \\
F_{v i}^{\prime}=Softmax\left(\frac{V_{i q}^T T_{i k}}{\sqrt{C_i}}\right) T_{i v}^T, \\
F_{v i}=W_{i o}\left(Unflatten\left(F_{v i}^{\prime T}\right)\right),
\end{gathered}
\end{equation}
Where $W_{iq}$, $W_{ik}$, $W_{iv}$ and $W_{iw}$ are learnable matrices.$\{F^{\prime}_{vi}\}_{i=2}^{4}\in \mathbb{R}^{H_{i} \times W_{i} \times C_i }$ are intermediate features. Eventually, we obtain the hierarchical multi-modal features $\{F_{vi}\}_{i=2}^{4}$ that incorporate linguistic information.

\begin{figure}[thbp]
  \centering
  \includegraphics[width=\linewidth]{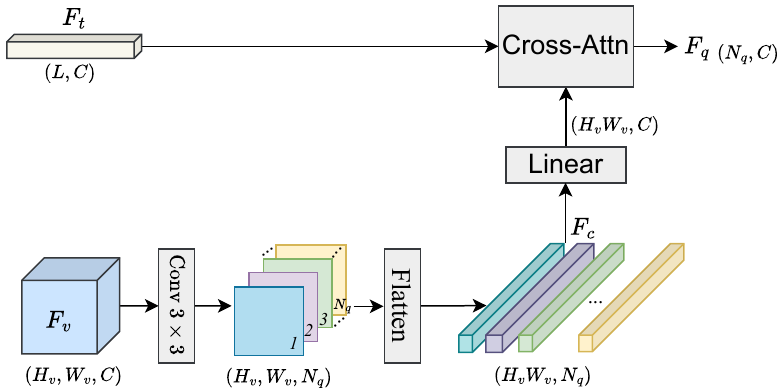}
  \caption{The details of Query Generation Module. }
  \vspace{0.0mm}
\end{figure}

\textbf{Fusion Neck}. In the Fusion Neck, we perform a straightforward fusion of hierarchical multi-modal features, including $F_{v 2}$, $F_{v 3}$, $F_{v 4}$. Initially, we upsample $F_{v 4}$ to obtain $F_{m 4} \in \mathbb{R}^{H_{3} \times W_{3} \times C}$ using the following equation:

\begin{equation}
F_{m 4}=U p\left(\sigma\left(F_{v 4} W_{v 4}\right) \right),
\end{equation}
In this process, $Up(\cdot)$ denotes 2 $\times$ upsampling function. Then We apply ReLU activation function which is denoted as $\sigma$ to generate $F_{m 4}$.
Subsequently, we obtain the multi-modal features $F_{m 3}$ and $F_{m 2}$ using the following procedures:
\begin{equation}
\begin{aligned}
& F_{m_3}=\left[\sigma\left(F_{m_4} W_{m_4}\right), \sigma\left(F_{v_3} W_{v_3}\right)\right], \\
& F_{m_2}=\left[\sigma\left(F_{m_3} W_{m_3}\right), \sigma\left(F_{v_2}^{\prime} W_{v_2}\right)\right], F_{v_2}^{\prime}=Avg\left(F_{v_2}\right),
\end{aligned}
\end{equation}
Where $Avg(\cdot)$ denotes a kernel size of 2 $\times$ 2 average pooling operation
, $[,]$ denotes the concatenation operation. Subsequently, we concatenate the three multi-modal features ($F_{m 4}, F_{m 3}, F_{m 2}$) and use a $1 \times 1$ convolution layer to aggregate them:
\begin{equation}
F_m= Conv\left(\left[F_{m_2}, F_{m_3}, F_{m_4}\right]\right),
\end{equation}
Where $F_{m} \in \mathbb{R}^{H_{3} \times W_{3} \times C}$. Then, we obtain the 2D spatial coordinate feature $F_{coord} \in \mathbb{R}^{H_{3} \times W_{3} \times 2}$ and concatenate it with $F_{m}$ and flatten the result to obtain the fused visual features with global textual information which is denoted as $F_{vt} \in \mathbb{R}^{H_{3}W_{3} \times C}$.
\begin{equation}
F_{v}=Conv\left(\left[F_m, F_{coord}\right]\right).
\end{equation}
Here, $Flatten(\cdot)$ denotes flatten operation and we obtain the $F_{v} \in \mathbb{R}^{H_v \times W_v \times C}$, $H_v = \frac{H}{16}$, and $W_v = \frac{W}{16}$.

\subsection{Language Query Generation}

 For referring image segmentation, the original language features extracted by the text encoder only contain one modality of information, which 
 the importance of different words in the same language expression is obviously different. Some previous works address this issue by measuring the importance of each word and giving each word a weight by the language self-attention. But what they neglect is that the importance of different words in the same language expression can vary depending on the specific image being referred to.  Therefore, we need to combine the language expression with the visual information to generate a set of language queries that are specific to the given image.  These queries contain more detailed language information than the original language features which only contain a single modality of information. Consequently, in the decoder process, these language queries are less likely to become distorted.

 In our approach, we use the multi-modal features $F_v$ and the language features $F_t$  to generate multiple queries, each corresponding to a different interpretation of the image. We do this by computing attention weights between the $F_v$ and $F_t$ for each query.

 As illustrated in Figure 4, the Query Generation Module takes $F_v$ and $F_t$ as input. We first utilize a $3\times3$ convolution to change the dimension of $F_v$ from $H_v\times W_v \times C$ to $H_v \times W_v \times N_q$. Here, $N_q$ serves as a hyperparameter representing the number of language queries. Then we flatten these features and obtain the intermediate features $F_c \in \mathbb{R}^{H_{v}W_{v} \times N_q}$:
 
\begin{equation}
F_c=Flatten(Conv(F_v)),
\end{equation}

 We utilize a cross-attention to generate queries. Firstly, we begin by applying linear projection to $F_{t}$ and $F_{c}$. Then, for the $n$-th query ($n = 1,2,...,N_q$), we take the $n$-th vision feature vector $f_{cn} \in \mathbb{R}^{1 \times \left(H_vW_v\right)}$. Specifically, we use $f_{ti} \in \mathbb{R}^{1 \times C}$ to denote the feature of the $i$-th word ($i = 1,2,...,L$). The attention weight for the $i$-th word with respect to the $n$-th query is computed as the product of projected $f_{vdn}$ and $f_{tvi}$:
\begin{equation}
a_{n i}=\sigma\left(f_{cn} W_{c}\right) \sigma\left(f_{ti} W_{t}\right)^T,
\end{equation}
In the equation, $a_{ni}$ represents a scalar that indicates the importance of the $i$-th word in the $n$-th query, where $W_{c}$ and $W_{t}$ are learnable matrices. To normalize the attention weights across all words for each query, we apply the Softmax function. The resulting values of $a_{ni}$, after being processed by Softmax, comprise the attention map $A \in \mathbb{R}^{N_q \times L}$. For the $n$-th query, we extract $A_n \in \mathbb{R}_{1 \times L}$ ($n = 1,2,...,N_q$) from A, which represents the emphasis of the words on the $n$-th query. And $A_n$ are use to generate the new queries as following equation:
\begin{equation}
    F_{qn} = A_n\sigma\left(F_{t} W_{q}\right). 
\end{equation} 
The matrix $W_{q}$ is a learnable matrix. The feature vector $F_{qn} \in \mathbb{R}^{1 \times C}$. Additionally, each new query is a projected weighted sum of the features of different words in the language expression. This enables the query to retain its properties as a language. $F_{qn}$ also represents a unique comprehension of the input sentence.  The set of all queries comprises the new language matrix $F_q \in \mathbb{R}^{N_q \times C}$, which is called the language query features.

 \begin{figure}[thbp]
  \centering
  \includegraphics[width=\linewidth]{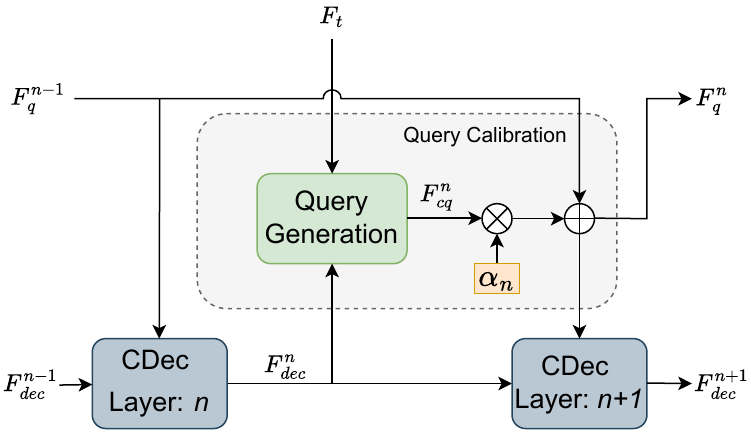}
  \caption{The architecture of one block of our proposed Calibration Decoder (CDec). }
  \vspace{0.0mm}
\end{figure}

\subsection{Calibration Decoder}
We employ a Calibration Decoder (CDec) to facilitate the transfer of fine-grained semantic information from textual features to visual features in an adaptive manner. As illustrated in Figure 2, the decoder takes query vectors $F_q$, language features $F_{t}$, and vision features $F_v$ as input. To incorporate positional information, we add $F_{t}$, $F_q$, and $F_v$ with sine spatial positional encodings. As illustrated in Figure 5, our CDec blocks include two main parts: a transformer decoder part and a Query Calibration part. 

\textbf{Transformer Decoder.} The transformer decoder part follows the standard transformer\cite{vaswani2017attention} design, where each layer consists of a multi-head self-attention layer, a multi-head cross-attention layer, and a feed-forward network. In the $n$-th decoder layer ($n = 1,2,..., N$), the multi-head self-attention layer is applied to $F_{dec}^{n-1}$, which is the output from the previous layer to capture global contextual information:
\begin{equation}
    F_{dec}^{\prime n} = MHSA\left(LN\left(F_{dec}^{n-1}\right)\right) + F_{dec}^{n-1},
\end{equation}
The resulting intermediate features are denoted as $F_{dec}^{\prime n}$, where $MHSA(\cdot)$ and $LN(\cdot)$ represent the multi-head self-attention layer and Layer Normalization\cite{ba2016layer}, respectively. In the first CDec layer, the input is vision features $F_v$. 

The multi-head self-attention mechanism consists of three linear layers that map $F_{dec}^{n-1}$ to intermediate representations, including queries $Q \in \mathbb{R}^{N \times d_q}$, keys $K \in \mathbb{R}^{N \times d_k}$, and values $V \in \mathbb{R}^{N \times d_v}$. 
Subsequently, we use a multi-head cross-attention layer to propagate fine-grained semantic information into the evolved visual features. Here, $Q$ is obtained by a linear projection of $F_{dec}^{\prime n}$, while $K$ and $V$ are both derived by two separate linear projections of $F_{q}^{n-1}$, which will be explained in the next paragraph. To obtain the multi-modal feature $F_s$, the output query $Q$ is processed through an MLP block comprising two layers with Layer Normalization and residual connections:

\begin{equation}
\begin{aligned}
&  F_{dec}^{\prime \prime n}=M H C A\left(L N\left(F_{dec}^{\prime n}\right), F_{q}^{n-1}\right)+F_{dec}^{\prime n}, \\
& F_{dec}^n=M L P\left(L N\left( F_{dec}^{\prime \prime n}\right)\right)+ F_{dec}^{\prime \prime n},
\end{aligned}
\end{equation}
Here, $MHCA(\cdot)$ denotes the multi-head cross-attention layer, and $F_{dec}^{\prime \prime n}$ represents the intermediate features. The output $F_{dec}^n$ is fed into the next CDec layer.

\textbf{Query Calibration.} 
In the traditional transformer, the language inputs remain the same with the depth of the layers. However, as the Q from the $F_{dec}{n}$ continues to update, fixed language input ${F_t}$ will cause information propagation to be gradually blocked as the number of layers increases. To address this issue, we utilize language queries as the language input and propose a Query Calibration part to adaptively calibrate the language queries based on each updated $F_{dec}^{n}$. 

As illustrated in Figure 5, the $n$-th Query Calibration part takes the language features $F_t$, the output of the  $n$-th CDec layer $F_{dec}^n$ and the language queries from the $(n-1)$-th Query Calibration part $F_q^{n-1}$ as input. We utilize a Query Generation Module to generate $F_{cq}^n$ to calibrate the language queries. $F_{cq}^n$ can be seen as the calibration queries. Here, the architecture of the Query Generation Module in the Query Calibration part is the same as the Query Generation Module described in Figure 4. The only difference is that we utilize $F_{dec}^{n}$ to replace the $F_v$ to generate queries based on the output of each layer. Then we utilize a learnable scalar $\alpha_n$ to control the calibration queries which are used to calibrate the $F_q^{n-1}$ by adding:

\begin{equation}
\begin{aligned}
&  F_{cq}^n = QGM(F_t,F_{dec}^n),\\
&  F_q^n = \alpha_nF_{cq}^n + F_q^{n-1},\\
\end{aligned}
\end{equation}

Here, QGM denotes the Query Generation Module. In the first Query Calibration part, the language query input is $F_q$. $F_q^n$ are the language queries incorporating the semantic information from this layer and will be input into the next Query Calibration part.

\begin{table*}[thbp]
    \setlength{\belowcaptionskip}{1.0pt}
    \begin{center}
    \caption{\textbf{Comparisons with the state-of-the-art approaches on three benchmarks.}
    We report the results of our method with various visual backbones.
    ``-'' represents that the result is not provided.
    mIoU is utilized as the metric.}
    \setlength{\tabcolsep}{2.8mm}{
    \begin{tabular}{l|c|ccc|ccc|cc}
        \toprule[1.2pt]
        \multirow{2}{*}{Method} & \multirow{2}{*}{Backbone} & \multicolumn{3}{c|}{RefCOCO} & \multicolumn{3}{c|}{RefCOCO+} & \multicolumn{2}{c}{G-Ref} \\
        \cline{3-10}
        ~ & ~ & val & test A & test B & val & test A & test B & val & test \\
        \midrule[1.2pt]
        MAttNet \cite{yu2018mattnet}           & ResNet-101 & 56.51 & 62.37 & 51.70 & 46.67 & 52.39 & 40.08 & 47.64 & 48.61 \\
        MCN \cite{luo2020multi}                & DarkNet-53 & 62.44 & 64.20 & 59.71 & 50.62 & 54.99 & 44.69 & 49.22 & 49.40 \\
        CGAN \cite{luo2020cascade}             & DarkNet-53 & 64.86 & 68.04 & 62.07 & 51.03 & 55.51 & 44.06 & 51.01 & 51.69 \\
        EFNet \cite{feng2021encoder}           & ResNet-101 & 62.76 & 65.69 & 59.67 & 51.50 & 55.24 & 43.01 & - & - \\
        LTS \cite{jing2021locate}              & DarkNet-53 & 65.43 & 67.76 & 63.08 & 54.21 & 58.32 & 48.02 & 54.40 & 54.25 \\
        VLT \cite{ding2021vision}                 & DarkNet-53 & 65.65 & 68.29 & 62.73 & 55.50 & 59.20 & 49.36 & 52.99 & 56.65 \\
        ReSTR \cite{kim2022restr}                    &  ViT-B &  67.22 & 69.30 & 64.45 & 55.78 & 60.44 &48.27 & 54.48 & - \\
        SeqTR \cite{li2021referring}                 & ResNet-101  & 70.56 & 73.49 & 66.57 & 61.08 & 64.69 & 52.73 & 58.73 & 58.51 \\
        CRIS \cite{wang2022cris}                           & ResNet-101 & 70.47 & 73.18 & 66.10 & 62.27 & 68.08 & 53.68 & 59.87 & 60.36 \\
        LAVT \cite{yang2022lavt}                           & Swin-Base  & 72.73 & 75.82 & 68.79 & 62.14 & 68.38 & 55.10 & 61.24 & 62.09 \\
         VLT+ \cite{ding2022vlt}                & Swin-Base& 72.96& 75.56& 69.60 &63.53& 68.43& 56.92& 63.49& 66.22\\
         RefSegformer \cite{wu2023towards}    & Swin-Base& 73.22& 75.64& 70.09 &63.50& 68.69& 55.44& 62.56& 63.07\\
        PVD \cite{cheng2023parallel} & Swin-Base&\underline{74.82}& \underline{77.11}& 69.52& 63.38 &68.60& 56.92& 63.13& 63.62\\
         $\rm{M^3}$AII \cite{liu2023multi}&  Swin-Base&73.60&76.23&\underline{70.36}&\underline{65.34}&\underline{70.50}&\underline{56.98}&\underline{64.92}&\textbf{67.37}\\
        \midrule
        CRFormer(Ours)        &Swin-Base &\textbf{75.26} &\textbf{ 77.38} & \textbf{71.92} & \textbf{66.98} & \textbf{71.74} & \textbf{59.32} & \textbf{65.97} & \underline{66.86}\\
    
        \bottomrule[1.2pt]
    \end{tabular}
    \label{tab:sota}}
    \end{center}
    \vspace{0.0mm}
\end{table*}

 \begin{figure}[thbp]
  \centering
  \includegraphics[width=0.85\linewidth]{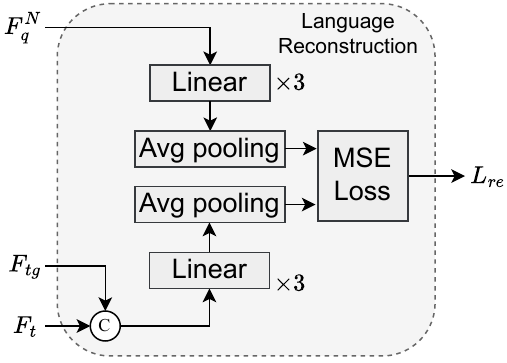}
  \caption{The process of our proposed Language Reconstruction. }
  \vspace{0.0mm}
\end{figure}

\subsection{Language Reconstruction}

In the preceding RIS models, the framework was solely optimized through the output mask loss. This implies that as long as the output mask aligns with the target region, it is presumed that the framework has grasped the input language and accomplished successful cross-modal semantic information propagation. However, in numerous instances, the network may fortuitously guess the correct target, either relying on distorted language information or randomly selecting a target, especially in scenarios where there are only a limited number of targets present in the image. In contrast to previous methods, we propose a novel language reconstruction loss to supervise the language information propagation. As illustrated in Figure 6, we utilize the language queries from the last decoder layer $F_q^{N} \in \mathbb{R}^{N_q\times C} $ to reconstruct the input language. So it ensures that the language information is well preserved through the whole multi-modal feature processing procedure.
\begin{equation}
    F_{re} = Avg(F_{q}^NW_{re}),
\end{equation}
Here, $W_{re} \in \mathbb{R}^{C \times C}$ represents three learnable projections and $Avg(\cdot)$ denotes average pooling on the first dimension.  $F_{re} \in \mathbb{R}^{1\times C}$ is the reconstructed language feature. Subsequently, we integrate the language features $F_t$ and the global language representation $F_t^\prime$, projecting them into a unified feature space through the following equation: 

\begin{equation}
F_{pt}=Avg (\sigma\left(\left[F_t,F_t^{\prime}\right] W_{pt }\right)) \text {, }
\end{equation}
Where $F_{pt}\in \mathbb{R}^{1\times C}$ denotes the projected language feature that represents the original language information. $W_{pt} \in \mathbb{R}^{C \times C}$ represents three learnable projections, $\sigma$ denotes ReLU function, [,] denotes concatenation. The language reconstruction loss is derived by minimizing the distance between the $F_{pt}$ and $F_{re}$ by mean squared error loss. For segmentation loss, we utilize a $3 \times 3$ convolution to obtain the segmentation mask, computing the segmentation loss by cross-entropy. The final loss function is defined as:

\begin{equation}
\mathcal{L}=\omega_{seg} \mathcal{L}_{seg}+\omega_{re} \mathcal{L}_{{re}},
\end{equation}
where $\omega_{seg}$ is the weight for segmentation loss $\mathcal{L}_{seg}$ and $\omega_{re}$ is the weight for  language reconstruction loss $\mathcal{L}_{re}$. The language reconstruction loss is computationally free during inference.

\section{Experiments}
\subsection{Implementation Details}
\textbf{Experiment Settings.} We strictly follow previous works \cite{yang2022lavt} for experiment settings. We utilize Swin-Base \cite{liu2021swin} as the vision encoder and BERT-Base \cite{devlin2018bert} as the text encoder, which is the same as the previous methods \cite{ding2022vlt,yang2022lavt,liu2023multi} 
Input images are resized to $480 \times 480$. The input sentences are set with a maximum sentence length of 20. Each transformer block has 8 heads, and the hidden layer size in all heads is set to 512, and the feed-forward hidden dimension is set to 2048. We train the network for 40 epochs using the Adam optimizer with the learning rate lr = 0.005. The weight for segmentation loss $\omega_{seg}$ is set to 1 and the language feature reconstruction loss $\omega_{re}$ is set to 0.1. We train the model with a batch size of 64 on 8 RTX Titan with 24 GPU VRAM.

\textbf{Metrics.} Following previous works \cite{wang2022cris,ding2021vision,yang2022lavt,liu2023multi}, we employ two metrics for evaluating effectiveness: Mean Intersection over Union (mIoU) and Precision@$\mathbf{X}$. IoU calculates the ratio of intersection regions to union regions between the predicted segmentation mask and the ground truth. The mIoU calculates the mean value of per-image IoU over all test samples. Precision@$\mathbf{X}$ gauges the percentage of test images with an IoU score surpassing the threshold $X \in {0.5, 0.6, 0.7, 0.8, 0.9}$, thereby emphasizing the method's spatial localization capability.
\subsection{Datasets}

We evaluate our method on three well-known datasets: RefCOCO \cite{yu2016modeling}, RefCOCO+ \cite{yu2016modeling}, and G-Ref \cite{nagaraja2016modeling}, all derived from the MS COCO dataset \cite{lin2014microsoft} and annotated with natural language expressions. RefCOCO contains 142,209 expressions for 50,000 objects in 19,992 images, while RefCOCO+ focuses on appearance expressions with 141,564 expressions for 49,856 objects in the same number of images. The primary distinction between the two lies in RefCOCO+ exclusively featuring appearance expressions, omitting words that denote location properties (e.g., left, top, front) in expressions. G-Ref contains 104,560 referring language expressions for 54,822 objects distributed across 26,711 images. Unlike RefCOCO and RefCOCO+, G-Ref exhibits a more casual yet complex language usage, with longer average sentence lengths. Additionally, the G-Ref dataset comprises two partitions: one established by UMD \cite{nagaraja2016modeling} and the other by Google \cite{mao2016generation}. In our paper, we present results specifically for the UMD partition.
\begin{figure}[htbp]
  \centering
  \includegraphics[width=0.95\linewidth]{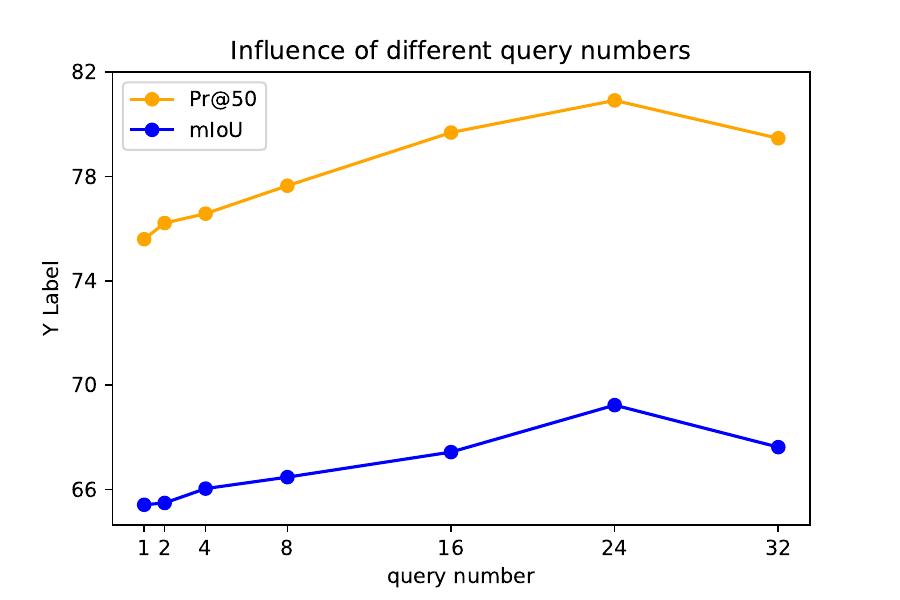}
  \caption{Performance of different query numbers $N_q$}
  \vspace{0.0mm}
\end{figure}
\begin{table}[htbp]
    \vspace{-0.0cm} 
    \setlength{\belowcaptionskip}{1.0pt}
    \begin{center}
    
    \caption{\textbf{Influence of Query Numbers.}}
    \setlength{\tabcolsep}{1.8mm}{
    \begin{tabular}{c|c|c|c|c|c|c}
        \toprule[1.2pt]
        $N_q$ & mIoU & Pr@50 & Pr@60 & Pr@70 & Pr@80 & Pr@90 \\
        \midrule
          32 & 67.62 & 79.46 & 75.15 & 68.76 & 53.42 & 17.50 \\
          24 & \textbf{69.23} & \textbf{80.91} & \textbf{77.45} & \textbf{71.46}  & \textbf{56.95}  & \textbf{18.11} \\
          16 & 67.43 & 79.68 & 75.79 & 68.73 & 52.84 & 14.85 \\
          8  & 66.65 & 77.64 & 73.09 & 66.07 & 49.17 & 15.01 \\ 
          4  & 66.03 & 76.57 & 72.58 & 65.45 & 48.95 & 14.67 \\
          2  & 65.48 & 76.21 & 72.39 & 64.32 & 48.52 & 14.02 \\
          1  & 64.61 & 75.69 & 71.58 & 63.47 & 47.69 & 14.16 \\
        \bottomrule[1.2pt]
    \end{tabular}
    \label{tab:Nq}}
    \end{center}
\end{table}

    

\subsection{Compare with others}
In Table 1, we evaluate CRFormer against the state-of-the-art referring image segmentation methods on the RefCOCO\cite{yu2016modeling}, RefCOCO+\cite{yu2016modeling}, and G-Ref\cite{nagaraja2016modeling} datasets using the mIoU metric.  Results show that our proposed method outperforms other methods on all three datasets. Compared to the second-best performing method $\rm{M^3}$AII \cite{liu2023multi}, our CRFormer model achieves absolute margins of 1.66, 1.15, and 1.56 scores on the validation, testA, and testB subsets of RefCOCO, respectively. Our proposed method also outperforms $\rm{M^3}$AII on the  RefCOCO+ dataset with 1.54, 1.24 and 2.34 absolute score improvements. On the G-Ref dataset, our method surpasses the second-best methods on the validation and test subsets from the UMD partition by absolute margins of 1.05 on the validation subset.

\subsection{Ablation Study}
We conduct several ablations to evaluate the effectiveness of the key components in our proposed network. we do the ablation study on the testA split of RefCOCO+, and the epoch is set to 25.

\textbf{Query Number}. In order to clarify the influence of the query number $N_q$, we set the $N_q$ to a series of different numbers. The results are reported at Table 2 and Figure 7. According to the result, multiple queries can improve the performance of our model which is about $4\%$ from 1 query to 24 queries.  However,  more $N_q$ is not always bring a better result, With the increase of $N_q$, the performance will gradually level off or even decline. Eventually, we choose $N_q=24$ for its best performance. 

\begin{table}[htbp]
    \begin{center}
    \caption{\textbf{Comprehensive comparison of  CDec and $\mathcal{L}_{{re}}$.}}
    \setlength{\tabcolsep}{1mm}{
    \begin{tabular}{c|c|c|c|c|c|c|c}
        \toprule[1.2pt]
        CDec&$\mathcal{L}_{{re}}$ & mIoU & Pr@50 & Pr@60 & Pr@70&Pr@80&Pr@90 \\
        \midrule
        \checkmark & \checkmark & \textbf{69.23} & \textbf{80.91} & \textbf{77.45} & \textbf{71.46}  & \textbf{56.95}  & \textbf{18.11} \\
        ~ & \checkmark & 67.76  &79.80 & 76.04 & 68.88 & 52.88 & 15.16 \\
        \checkmark & ~ & 67.98 & 80.03 & 76.27 & 69.55 & 53.37 & 15.43 \\

            ~ & ~ & 66.93 & 78.23 & 74.27 & 68.45 & 51.76 & 14.43\\
        \bottomrule[1.2pt]
    \end{tabular}
    \label{tab:Nq}}
    \end{center}
\end{table}

\begin{table}[htbp]
    \begin{center}
    \caption{\textbf{Comparison of  CDec and standard transformer decoder with different decoder layers $N$.}}
    \setlength{\tabcolsep}{1.2mm}{
    \begin{tabular}{c|c|c|c|c|c|c|c}
        \toprule[1.2pt]
        $N$ & CDec &mIoU & Pr@50 & Pr@60 & Pr@70 & Pr@80 & Pr@90 \\
        \midrule
         \multirow{2}{*}{4} & \checkmark & 68.45 & 80.37 & 76.92 & 70.16 & 54.07 & 16.34 \\
          \cline{2-2}
          ~ & & 67.14 & 78.84 & 75.58 & 68.51 & 52.56 & 14.53 \\
          \midrule
          \multirow{2}{*}{3} & \checkmark & \textbf{69.23}   &\textbf{80.91}  & \textbf{77.65} & \textbf{71.46} & \textbf{56.45} & \textbf{18.11} \\
            \cline{2-2}
          ~ & & 67.76  &79.80 & 76.04 & 68.88 & 52.88 & 15.16 \\
          \midrule
          \multirow{2}{*}{2} & \checkmark & 68.37 & 80.41 & 77.24 & 70.03 & 53.91 & 16.12 \\
          \cline{2-2}
          ~ & & 67.25 & 78.53 & 74.95 & 68.61 & 52.31 & 14.67 \\
          \midrule
          \multirow{2}{*}{1} & \checkmark & 66.69 & 78.19 & 74.21 & 67.94 & 51.65 & 14.24 \\
          \cline{2-2}
          ~ & & 66.12 & 77.28 & 73.57 & 67.05 & 51.10 & 13.51 \\
 
        \bottomrule[1.2pt]
    \end{tabular}
    \label{tab:Nq}}
    \end{center}
    \vspace{-0.0mm}
\end{table}
\begin{table}[htbp]
    \vspace{-0.0cm} 
    \setlength{\belowcaptionskip}{1.0pt}
    \begin{center}
    
    \caption{\textbf{Ablation study of loss function.}}
    \setlength{\tabcolsep}{1.8mm}{
    \begin{tabular}{c|c|c|c|c|c|c}
        \toprule[1.2pt]
        $\omega_{re}$ & mIoU & Pr@50 & Pr@60 & Pr@70 & Pr@80 & Pr@90 \\
        \midrule
            0.20  & 66.47 & 77.64 & 73.09 & 66.07 & 49.17 & 15.01 \\
          0.15 & 67.43 & 79.68 & 75.79 & 68.73 & 52.84 & 14.85 \\

          0.10 & \textbf{69.23} & \textbf{80.91} & \textbf{77.45} & \textbf{71.46}  & \textbf{56.95}  & \textbf{18.11} \\
          0.05  & 68.38 & 80.21 & 76.57 & 70.06 & 54.25 & 16.14 \\
          0.00  & 67.98 & 80.03 & 76.27 & 69.55 & 53.37 & 15.43 \\
        \bottomrule[1.2pt]
    \end{tabular}
    \label{tab:Nq}}
    \end{center}
\end{table}

\begin{figure*}[thbp]
  \centering
  \includegraphics[width=\linewidth]{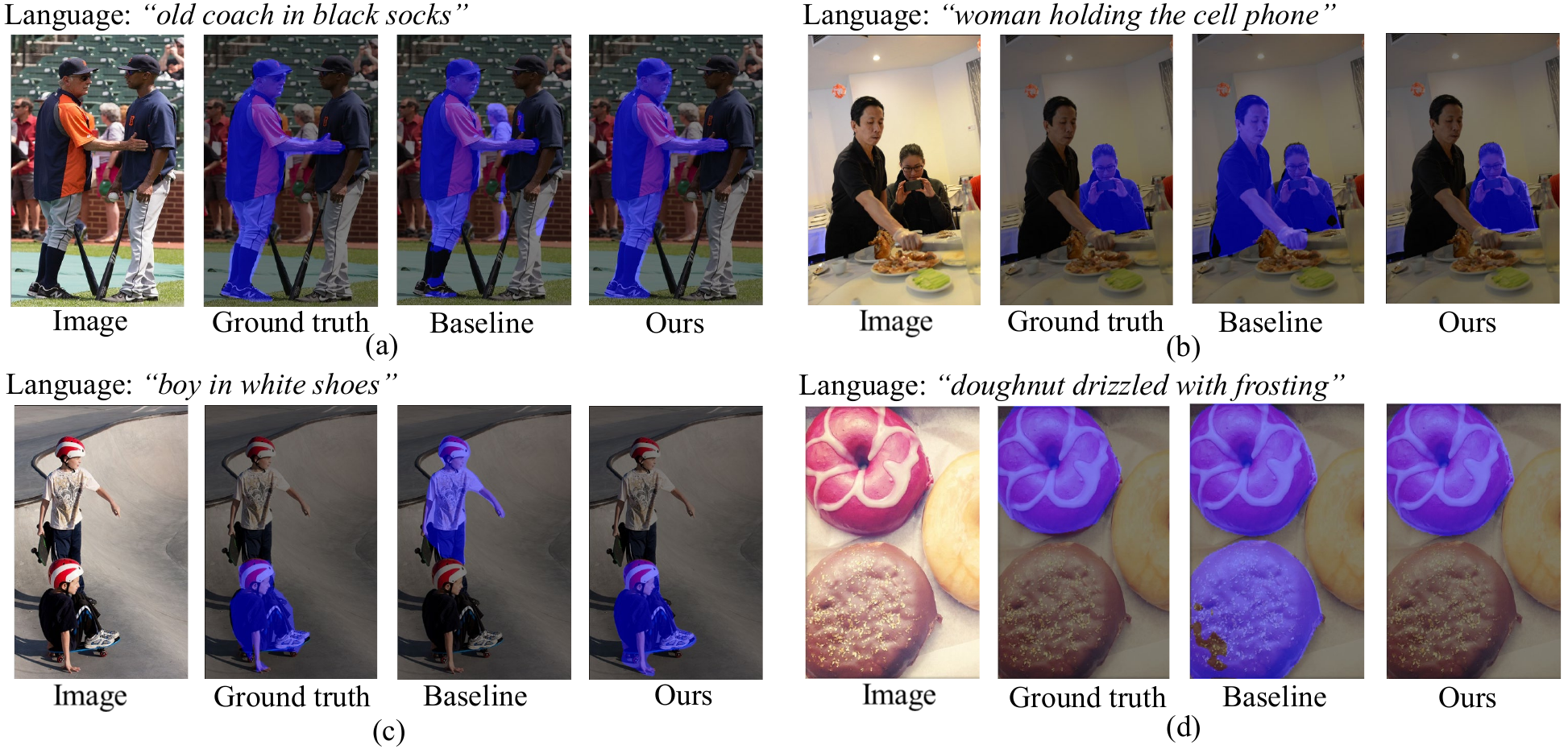}
  \caption{Qualitative examples. (a) the input image. (b) the ground truth. (c) the baseline network which utilizes the standard transformer decoder. (d) our proposed CRFormer }
\end{figure*}

\textbf{Comprehensive Comparison}. We propose two main components to address language distortion: the Calibration Decoder (CDec) and a novel language reconstruction loss $\mathcal{L}{re}$. To comprehensively evaluate their effectiveness, we present a comparison of each component. Detailed ablations of each part are provided in the following paragraphs. As illustrated in Table 3, removing the CDec results in a decrease of 1.47; removing $\mathcal{L}{re}$ leads to a decrease of 1.25; removing both components yields an even more inferior result, with a decrease of 2.30. This comprehensive comparison validates the significance of each component in the propagation process, and adopting both together can synergistically enhance results.

\textbf{Calibration Decoder}. To substantiate the efficacy of our proposed Calibration Decoder (CDec), we perform comparative experiments by substituting the CDec with a conventional transformer decoder. Additionally, we present the results for different layers $N$ in both of them. As illustrated in Table 4, CDec outperforms the standard transformer across various settings of decoder layers, which significantly increases the mIoU accuracy of $2.0\%, 2.2\%, 1.7\%$ and $0.9\%$. This superior performance gain proves that the CDec can prevent language information distortion which is vital for cross-modal semantic information propagation. Furthermore, it is noteworthy
that the optimal performance is achieved when $N = 3$. The setting
of $N = 1$ may not take full advantage of the multi-modal corresponding information from both vision and language. The setting
of $N = 4$ may increase the risk of over-fitting. Therefore, we choose $N=3$ as the default for its performance. 

\textbf{The loss function}. Our loss function comprise two parts: segmentation loss $\mathcal{L}_{seg}$ and language reconstruction loss  $\mathcal{L}_{re}$, which are both weighted by two parameters: $\omega_{seg}$ and $\omega_{re}$. In order to clarify the importance of $\mathcal{L}_{re}$, we set $\omega_{seg}$ to 1 and choose different $\omega_{re}$ from 0 to 0.2. As illustrated in Table 5, we present the results corresponding to various settings of $\omega_{re}$. When $\omega_{re}=0$, the total loss function comprises solely the segmentation loss $\mathcal{L}{seg}$. This leads to a reduction of 1.25 in mIoU compared to $\omega{re}=0.10$. As $\omega_{re}$ increases further, the results deteriorate, indicating that the model overly emphasizes the reconstruction of language information at the expense of the accuracy of image segmentation results. Finally, we select $\omega_{re}=0.10$ as the default setting due to its superior performance.

\section{Visualization}
As shown in Figure 8, we present visualization results to showcase the effectiveness of our proposed CDec. In our baseline approach, a standard transformer decoder is employed. A comparison with our method reveals that the baseline fails to effectively propagate semantic information from language to image, resulting in inferior outcomes. For instance, in Figure 8(c), the input sentence is "boy in white shoes," yet in the baseline model, the "shoes" are distorted, leading to the model outputting a boy in a white T-shirt instead. Similarly, in Figure 8(d), the input sentence "doughnut drizzled with frosting" is incorrectly rendered by the baseline model, which outputs both doughnuts with distorted language information. In contrast, our model successfully predicts the correct mask, as it retains the crucial language information "drizzled with frosting."

\section{Conclusion}
In this paper, we propose an end-to-end framework, CRFormer, which gradually updates the language features and deep integrates the language to address the problem of language information distortion during the semantic propagation process in referring image segmentation. We address this problem in the following three ways. Firstly, We generate multiple language queries that represent different emphases of the input so that the key language information can be retained. Secondly, we design a Calibrated Decoder to gradually update and calibrate the language input by generating new language queries. Then, we employ a language reconstruction loss which supervises the language propagation process. Our experiments show that our method significantly outperforms previous state-of-the-art methods on RefCOCO, RefCOCO+ and G-Ref datasets without any post-processing. 

\section*{Acknowledgments}
This work was supported by the National Science and Technology Major Project (No.2022ZD0118801), National Natural Science Foundation of China (U21B2043, 62206279).
\balance
\printbibliography


\end{document}